\documentclass[lettersize,journal]{IEEEtran}
\usepackage{amsmath,amsfonts}
\usepackage{algorithmic}
\usepackage{algorithm}
\usepackage{array}
\usepackage{booktabs}
\usepackage[caption=false,font=normalsize,labelfont=sf,textfont=sf]{subfig}
\usepackage{textcomp}
\usepackage{stfloats}
\usepackage{url}
\usepackage{verbatim}
\usepackage{graphicx}
\usepackage{cite}
\usepackage[numbers, sort&compress]{natbib}
\usepackage{enumitem}
\usepackage{multirow}
\usepackage{makecell}
\usepackage[table]{xcolor}
\usepackage{float}
\usepackage{caption}
\usepackage{hyperref}

\begin{document}

\title{RoadFormer : Local-Global Feature Fusion for Road Surface Classification in Autonomous Driving}

\author{Tianze Wang*, Zhang Zhang*, Chao Sun
    \thanks{Tianze Wang, Chao Sun and Zhang Zhang are with Beijing Institute of Technology.}
    }

\markboth{Journal of \LaTeX\ Class Files,~Vol.~14, No.~8, August~2025}%
{Shell \MakeLowercase{\textit{et al.}}: A Sample Article Using IEEEtran.cls for IEEE Journals}



\twocolumn[{
\renewcommand\twocolumn[1][]{#1}
\maketitle
\vspace{-0.8cm} 
\begin{center}
    \captionsetup{type=figure}
    \includegraphics[width=2.1\columnwidth]{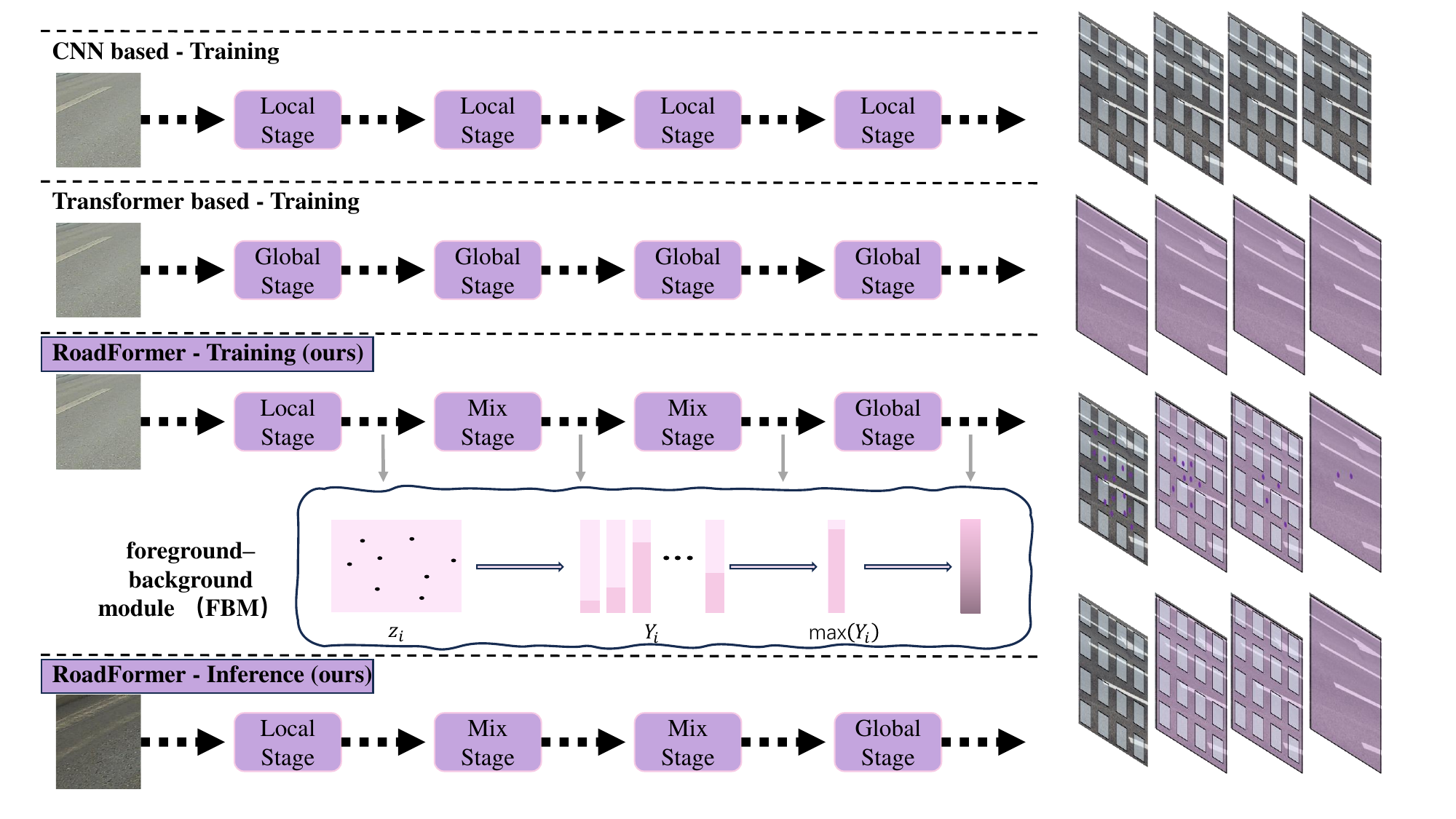}
    \captionof{figure}{\textbf{RoadFormer structural diagram.} Compared to CNN-based methods and Transformer-based methods, our proposed RoadFormer combines local and global features through a novel stacking structure and introduces a foreground-background module in the training process to extract image regions beneficial for classification tasks.}
    \label{jian_tu}
\end{center}
}]

\footnotetext{
* Equal contribution

Tianze Wang, Chao Sun and Zhang Zhang the National Engineering Laboratory for Electric Vehicles, School of Mechanical Engineering, Beijing Institute of Technology, Beijing 100081, China.}

\begin{abstract}
The classification of the type of road surface (RSC) aims to utilize pavement features to identify the roughness, wet and dry conditions, and material information of the road surface. Due to its ability to effectively enhance road safety and traffic management, it has received widespread attention in recent years. In autonomous driving, accurate RSC allows vehicles to better understand the road environment, adjust driving strategies, and ensure a safer and more efficient driving experience. For a long time, vision-based RSC has been favored. However, existing visual classification methods have overlooked the exploration of fine-grained classification of pavement types (such as similar pavement textures). In this work, we propose a pure vision-based fine-grained RSC method for autonomous driving scenarios, which fuses local and global feature information through the stacking of convolutional and transformer modules. We further explore the stacking strategies of local and global feature extraction modules to find the optimal feature extraction strategy. In addition, since fine-grained tasks also face the challenge of relatively large intra-class differences and relatively small inter-class differences, we propose a Foreground-Background Module (FBM) that effectively extracts fine-grained context features of the pavement, enhancing the classification ability for complex pavements. Experiments conducted on a large-scale pavement dataset containing one million samples and a simplified dataset reorganized from this dataset achieved Top-1 classification accuracies of 92.52\% and 96.50\%, respectively, improving by 5.69\% to 12.84\% compared to SOTA methods. These results demonstrate that RoadFormer outperforms existing methods in RSC tasks, providing significant progress in improving the reliability of pavement perception in autonomous driving systems.
\end{abstract}


\section{Introduction}
Autonomous vehicle technology is evolving rapidly, and many companies are developing systems for various vehicle types. In recent years, the automotive industry has implemented partial automation systems, while fully autonomous driving systems are still in advanced testing. Meanwhile, with the rapid development of urban infrastructure and growing demand for intelligent transportation, timely and accurate road surface recognition and classification have become crucial, directly impacting road maintenance, traffic management, and autonomous driving systems.

\begin{figure}[t]
    \centering
    \includegraphics[width=0.5\textwidth]{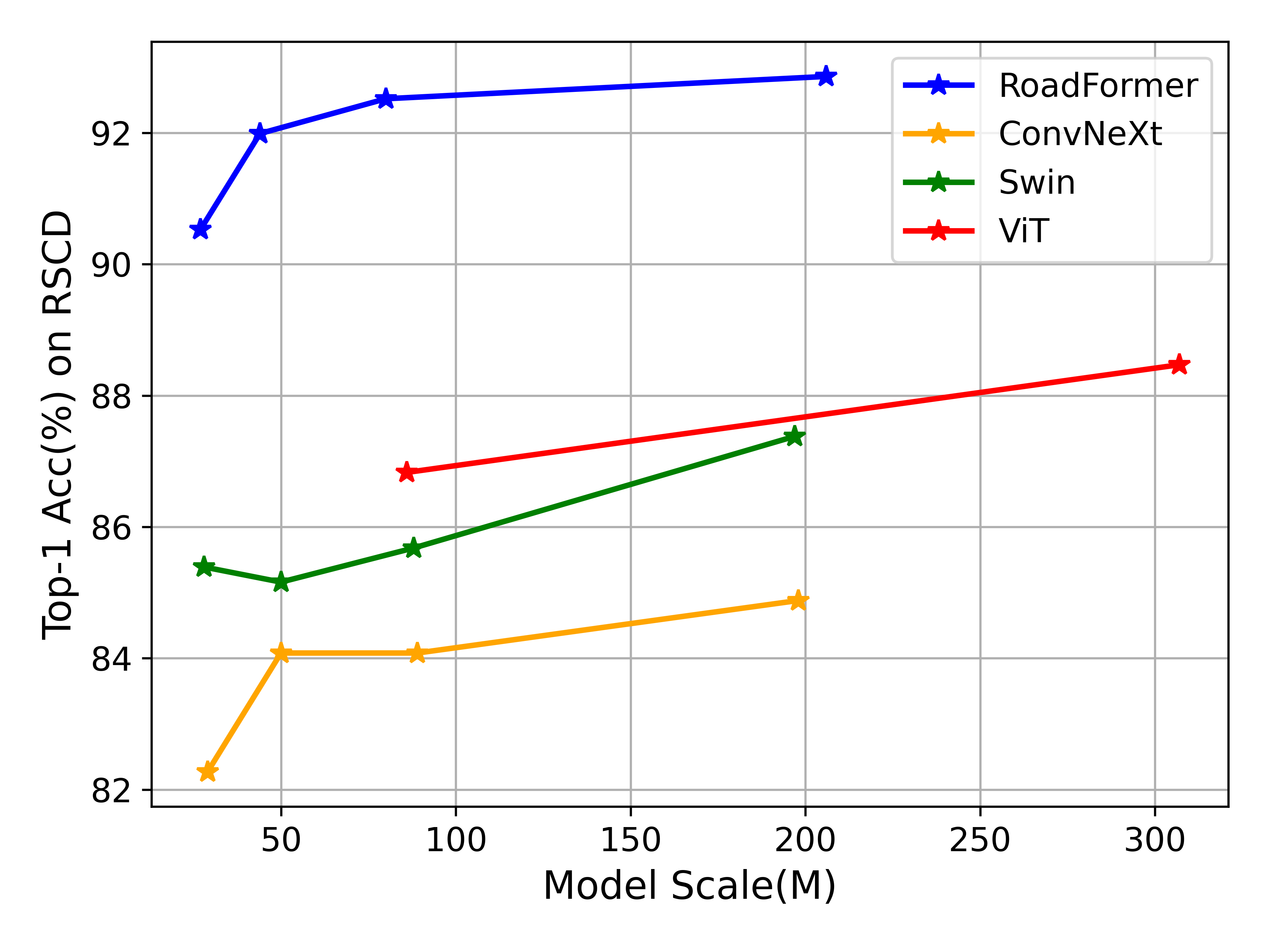}
    \caption{\textbf{Comparison among RoadFormer and efficient Networks.} The horizontal axis represents the number of model parameters, and the vertical axis represents the comparison metrics, namely Top-1 Acc.}
    \label{sota_curve}
\end{figure}

In this context, Road Surface Classification (RSC) is crucial. By discerning road friction, material, and unevenness, autonomous vehicles can optimize driving modes and stability control to adapt to various conditions, ensuring safety and comfort. After RSC, vehicles make key decisions for safe operation. They assess risk based on road conditions and adjust speed; for instance, slowing down on severely damaged or slippery roads. Also, vehicles may change trajectories to dodge danger, like making steering adjustments or re-planning routes in harsh conditions. Moreover, the system can activate safety protocols, heightening brake sensitivity and stability control. These real-time decisions protect the autonomous driving system and passenger safety.

Existing RSC methods include traditional and deep learning-based ones. Traditional methods split into vibration response-based, using accelerometers and displacement sensors to collect system responses from pavement excitations for classification \cite{1, 2}, and pavement feature-based, leveraging vehicle-mounted cameras and traditional image processing to extract texture or color features for classification \cite{3, 4, 5, 6}. Some combined both, like Bekhti et al. \cite{7}, who first captured pavement images, estimated texture characteristics, and correlated them with vibration to predict pavement conditions. However, these traditional methods suffer from high labor costs, limited coverage, and accuracy issues. For instance, on rough non-urban roads, accelerometers and displacement sensors struggle to ensure identification accuracy, and pavement texture changes over time degrade performance \cite{8, 9, 10}. In contrast, deep learning has revolutionized the field. As a machine learning branch excelling in image recognition tasks, it benefits from the ubiquity of cameras in intelligent vehicles. Vision-based deep learning for pavement perception has proven effective \cite{11} and is widely used to enhance accuracy \cite{12, 13, 14, 66} or efficiency \cite{15, 67, 68} in road surface classification and risk detection.
\IEEEpubidadjcol

However, the application of deep learning methods in RSC still faces many challenges. Firstly, existing deep learning models in classification tasks only focus on local features or global features, resulting in bias. To address this, we propose a framework that cross-extracts local and global features and further explore the stacking strategy within the framework. In addition, RSC, as a fine-grained classification task, faces the challenge of relatively large intra-class differences and relatively small inter-class differences, for which we introduce the Foreground-Background Module (FBM) to improve. In summary, we propose a method for fine-grained road surface classification aimed at autonomous driving scenarios, called RoadFormer, which enhances the classification ability of complex fine-grained tasks through novel stacked local-global feature extraction modules and FBM, ensuring the safety, comfort, and controllability of autonomous driving. Our main contributions are as follows:
\begin{itemize}
    \item We propose a novel hybrid convolution-transformer backbone, efficiently combines local and global feature fusion, allowing for the extraction of detailed road surface textures while also capturing global connections between pixels. This dual-level feature extraction enables autonomous vehicles to more accurately interpret road surfaces.
    \item Additionally, we explore advanced stacking strategies for convolution and transformer layers, providing a more robust architecture for road surface classification under real-world conditions, where variable texture similarities pose significant challenges.
    \item Furthermore, this study introduces a foreground-background separation module to address the challenges of large intra-class sample differences and small inter-class sample differences in fine-grained road surface classification.

\end{itemize}

\section{Related Work}
\textbf{CNN-based methods.} The release of the LeNet5 model \cite{16} in 1998 marked the true emergence of CNNs, defining their basic structure. In 2012, AlexNet's victory in the ImageNet challenge \cite{17} was a milestone in computer vision. Subsequently, scholars optimized CNN models: VGG \cite{18} deepened the network, GoogLeNet \cite{19} introduced the Inception module, ResNet \cite{20} solved the vanishing gradient problem, DenseNet \cite{21} enhanced information flow, and SENet \cite{22} incorporated an attention mechanism. CNN-based methods have excelled in visual tasks like classification, detection, and segmentation.

In RSC research, some works focused on model application and improvement. Using the pretrained VGG16, a study \cite{23} analyzed a Canadian highway dataset for winter road conditions. Roychowdhury et al. \cite{24} proposed a two - stage method for indirect pavement friction estimation. Cheng et al. \cite{25} introduced Gai - ReLU to improve classification accuracy. RCNet \cite{26} proposed a CNN-based model for classifying roads into five categories. Carrillo et al. evaluated several SOTA models (InceptionV3, Xception, MobileNetV2, NASNet) and developed a simplified baseline model \cite{27, 28, 29, 30, 31}.

\textbf{Transformer-based approach.} Transformer \cite{32} is a sequence modeling method based on the self-attention mechanism, initially developed for natural language processing and later widely applied in the field of computer vision. ViT \cite{33} represents the first application of Transformer in computer vision, transforming the image classification problem into a sequence modeling problem. Subsequently, the Swin Transformer \cite{34} introduced a sliding window-based self-attention mechanism, combining a local receptive field, which improved both computational efficiency and accuracy. This model has achieved significant performance improvements on various computer vision tasks. Swin Transformer V2 \cite{35} further optimized the structure of the original Swin Transformer, improving the performance of the model and the stability of the training.

Leveraging the powerful long-range dependency capture ability of Transformer models, Lin et al. \cite{15} proposed an anomaly detection method for road surfaces based on Transformers and self-supervised learning. Samo et al. \cite{37} used ViT to address road-specific tasks for weather recognition, utilizing focal loss to significantly improve the accuracy of computer vision methods. Furthermore, Transformer-based approaches have demonstrated remarkable performance in other image-classification tasks. DeiT \cite{38}, which is based on the Transformer architecture, achieves efficient image classification with fewer parameters, making it suitable for data-limited scenarios. Through knowledge distillation and small model parameters, the algorithm achieves satisfactory performance with fewer parameters. However, when facing large-scale datasets, models often encounter convergence and performance issues as network depth increases. To address this, CaiT \cite{39} introduced LayerScale and Class Attention, which significantly improved the accuracy and training effectiveness of deep models. T2T-ViT \cite{40} improved the model's ability to capture fine details by introducing a token-to-token module, which improved the expressive power of the image patches.

\textbf{CNN-Transformer hybrid architecture.} Hybrid CNN-Transformer models have become a prominent research topic in computer vision. CNNs excel in capturing local features, while Transformers are better suited for global contexts, but struggle with local details. Integrating both models enables effective processing of both local and global information, enhancing performance. Various integration strategies include novel architectural designs, series-parallel concatenations, and local substitutions.

In architectural designs, some models \cite{41} enhance ViT by incorporating CNN-inspired pyramid structures for improved high-resolution image processing. Others, like UNeXt \cite{42} and Uformer \cite{43}, draw from the UNet \cite{44} architecture to optimize Transformer use in vision tasks. The CSWin Transformer \cite{45} incorporates residual-like connections in a multi-layer Transformer structure, while HRFormer \cite{46} integrates Transformer modules into HRNet’s multi-resolution feature fusion, improving long-range dependency modeling and global semantic information.

In series-parallel concatenation, Carion et al. \cite{49} proposed DETR, where CNN extracts 2D features and reshapes them into feature sequences for the Transformer to perform object detection. This reduces the input size for faster learning of global features. Beal et al. \cite{50} introduced ViT-FRCNN, concatenating Faster R-CNN after ViT to use Transformer for object detection. Unlike this, Peng et al. \cite{52} used parallel concatenation in Conformer, fusing CNN’s local features and Transformer’s global features via a Feature Coupling Unit. Chen et al. \cite{53} proposed Mobile-Former, using a bidirectional cross-bridging method to integrate local and global features.

For local replacement, ViTC \cite{54} replaces 16×16 convolutions in ViT with stacked smaller convolutions, improving performance on ImageNet-1k. LocalViT \cite{55} introduces depth convolution into ViT's feed-forward network to add locality, while ConViT \cite{56} substitutes the self-attention layer with a Gated Positional Self-Attention (GPSA) layer, enabling controllable inductive bias based on contextual information.

\textbf{Fine-grained classification.} Fine-grained image classification focuses on recognizing subcategories within broader categories and is widely used in both industrial and academic fields. However, it is challenging due to subtle differences between subcategories, large intra-class variations, and factors like viewpoint, background, and occlusion.

Deep learning methods have become central to fine-grained image classification, yielding promising results. Zhang et al. \cite{57} introduced the Picking Deep Filter Responses framework, which extracts deep filter responses through a two-step process: first, identifying filters with consistent responses to specific patterns and learning part detectors, and second, aggregating responses using a spatially weighted combination of Fisher vectors. Zheng et al. \cite{58} addressed the gap in local localization and feature learning by proposing a multi-attention CNN model, which enhances both local region learning and feature extraction. Another approach divides the fine-grained dataset into visually similar subsets, as in the subset feature learning network \cite{59}, which combines a domain-general CNN pre-trained on a large dataset and several specific CNNs fine-tuned on the target dataset. Additionally, Wang et al. \cite{60} proposed the Multi-granularity CNN, a parallel deep CNN model for classification at various granularities, leveraging hierarchical subcategory labels for discriminative region extraction.

\begin{figure*}[t]
    \centering
    \includegraphics[width=\textwidth]{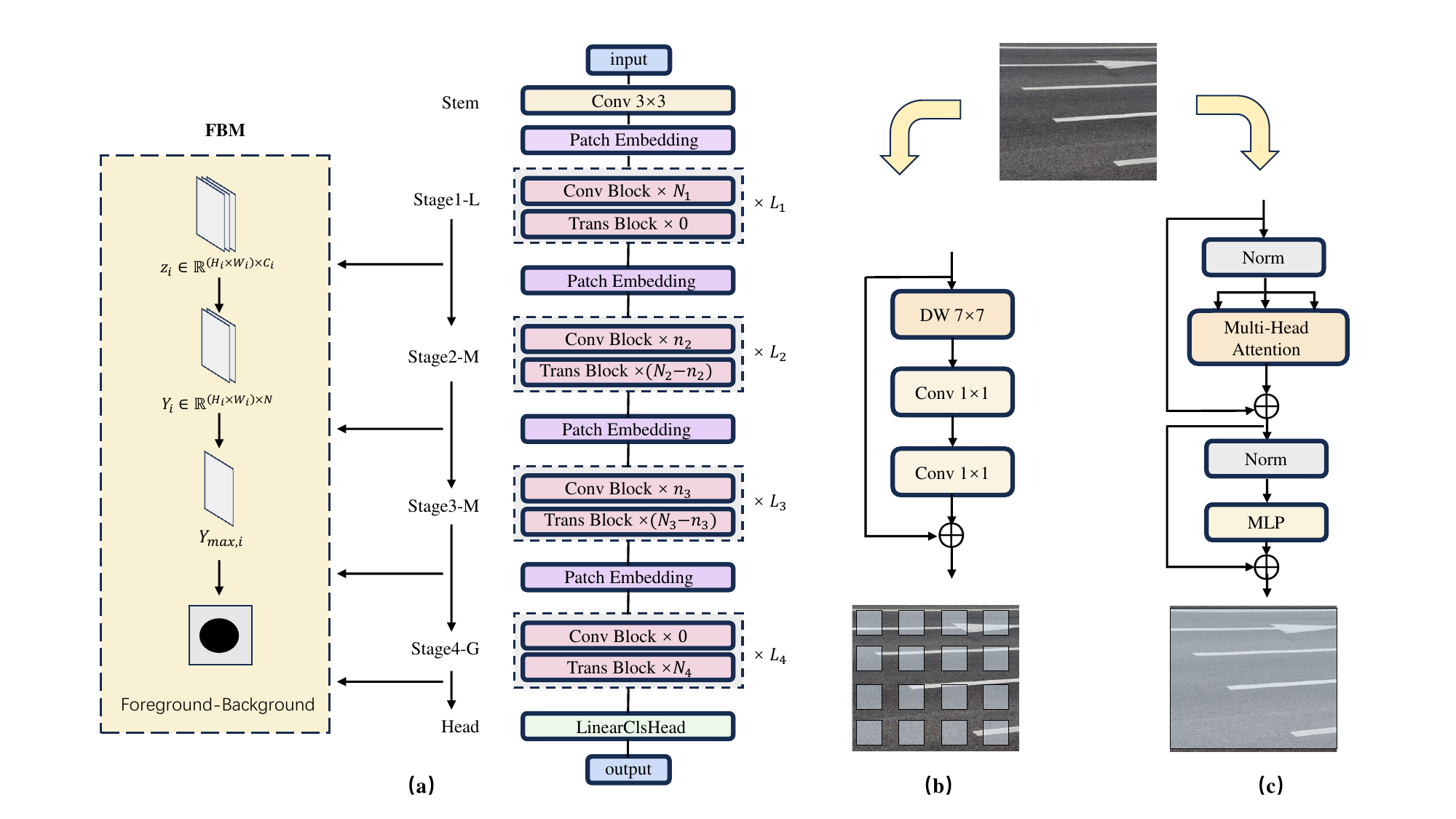}
    \caption{\textbf{RoadFormer Network Details.} (a) RoadFormer detailed network architecture. The first stage is the Local stage, using only the Conv Block. The second and third stages are the Mix stages, flexibly stacking the Conv Block and Trans Block. The fourth stage is the Global stage, using only the Trans Block. (b) Conv Block. Stacked by one layer of Depthwise convolution and two layers of $1\times1$ convolution. (c) Trans Block. Stacked by MHA and MLP. }
    \label{RoadFormer}
\end{figure*}

\section{Methods}

In this section, we first present the overall architecture of the proposed RoadFormer. Then, we discuss the foreground-background Module tailored for fine-grained classification tasks. Additionally, we provide the architectural specifications for different model sizes.

\textbf{Overall Architecture.} In order to achieve efficient and accurate road surface classification, we designed the RoadFormer network. As shown in Figure~\ref{RoadFormer}, the entire network is constructed from the bottom up with four main stages, aiming to couple local convolutional representations with global attention mechanisms through a novel stacking approach to enhance fine-grained classification performance.

First, the input image is passed through a $3\times3$ convolutional layer and subsequent Patch Embedding units to generate the initial features, this part is called the Stem module. Then, the initial features sequentially enter four stages (Stage1-L, Stage2-M, Stage3-M and Stage4-G), where L represents the local feature extraction stage, M represents the mixed feature extraction stage, and G represents the global feature extraction stage. The key to each stage lies in the stacking method of the Conv Block and the Trans Block. It should be noted that before entering each stage, feature downsampling needs to be carried out through the Patch Embedding layer. In Stage1-L, only the Conv Block is repeatedly stacked $N_1$ times, focusing on local context capture. In Stage2-M and Stage3-M, the Conv Block and the Trans Block are stacked according to the specified number of times respectively, so as to achieve the interaction of local and global features at different resolutions. After entering Stage4-G, the modeling ability to strengthen global dependencies at a deeper level is realized by retaining the Trans Block $N_4$ times only. As an internal component of the entire model, the Conv Block adopts a combined structure of depthwise separable convolution and 1×1 convolution. It obtains stable and efficient local representations through residual connections and feature channel recombination. The Trans Block, on the other hand, utilizes multi-head attention and MLP layers, and enhances the global perception of features with the assistance of normalization and skip connections. Finally, the high-level semantic information from Stage4-G is outputted as prediction results through the classification head. In general, this hierarchical stacking design takes into account both local and global information and has good scalability and transferability, providing a powerful foundation for feature representation for subsequent multiscale object detection and recognition tasks.

\textbf{Foreground Background Module (FBM).} A feature difference enhancement module, which we call the foreground-background module, runs through the model structure. After each feature extraction stage, the foreground-background separation operation is carried out through this module. 
The aim is to focus on the foreground features to enhance the category discrimination ability and to suppress the background information to reduce the influence of irrelevant features on the classification task. The feature map output at each stage is denoted as $z_i \in \mathbb{R}^{(H_i \times W_i) \times C_i}, i \in [1,2,3,4]$, where $i$ is the code for each feature extraction stage. Based on the feature map of each stage, FBM constructs the corresponding classification map, as shown in Eq. (\ref{yi}):
\begin{equation}
  Y_i = \text{Softmax}(w_i z_i + b_i) \label{yi}
\end{equation}
where $w_i$ represents the weight of the classifier at the $i$-th stage, and $b_i$ is its bias. $Y_i$ is the classification map with a size of $\mathbb{R}^{(H_i \times W_i) \times C}$, and \(N\) is the number of target categories. 

\begin{table*}[htbp]
    \centering
    \caption{Detailed configurations of RoadForme variants.} 
        \renewcommand{\arraystretch}{1.5}
        \begin{tabular}{ccc|c|c|c}
           \toprule
           Stages & Layers & RoadFormer-T & RoadFormer-S & RoadFormer-B & RoadFormer-L \\
           \midrule
           \multirow{2}*{Stage 1} & Patch Embedding & \multicolumn{4}{c}{Conv $3\times3$, $C=96$} \\
           \cmidrule(lr){2-6}
           & Local Block & \multicolumn{4}{c}{[Conv Block $\times 3$]} \\
           \midrule
           \multirow{2}*{Stage 2} & Patch Embedding & \multicolumn{4}{c}{Conv $3\times3$, $C=192$} \\
           \cmidrule(lr){2-6}
           & Mix Block & \multicolumn{4}{c}{[Conv Block $\times 3$ + Trans Block $\times 1$]} \\
           \midrule
           \multirow{2}*{Stage 3} & Patch Embedding & \multicolumn{4}{c}{Conv $3\times3$, $C=384$} \\
           \cmidrule(lr){2-6}
           & Mix Block 
           & \makecell[c]{[Conv Block $\times 3$ + \\ Trans Block $\times 2$] $\times 1$} 
           & \makecell[c]{[Conv Block $\times 3$ + \\ Trans Block $\times 2$] $\times 2$}
           & \makecell[c]{[Conv Block $\times 3$ + \\ Trans Block $\times 2$] $\times 3$}
           & \makecell[c]{[Conv Block $\times 3$ + \\ Trans Block $\times 2$] $\times 4$} \\
           \midrule
           \multirow{2}*{Stage 4} & Patch Embedding & \multicolumn{4}{c}{Conv $3\times3$, $C=768$} \\
           \cmidrule(lr){2-6}
           & Global Block & [Trans Block $\times 2$] & \multicolumn{3}{c}{[Trans Block $\times 3$]} \\
           \midrule
           \multicolumn{2}{c}{Output Channel} & 768 & 768 & 1024 & 1536 \\
           \bottomrule
        \end{tabular} \label{variants}
\end{table*}

This module contains a selector, which is used to distinguish between the foreground and background regions from the classification map (in RSC task, the foreground region refers to the area that contains rich pavement-type features). Calculate the maximum score map at the $i$-th stage according to the classification map, as shown in Eq. (\ref{ymax}):

\begin{equation}
  Y_{\text{max},i} = \max(Y_i) \label{ymax}
\end{equation}

Next, the obtained $Y_{\text{max},i}$ is sorted in descending order, and the top $K_i$ features $Y_{\text{max},i}^{K_i}$ with the highest scores are selected as the foreground features, while the other features $Y_{\text{max},i}^{\overline{{K_i}}}$ are regarded as the background features.  Considering that the processing of the subsequent network is carried out based on the output of the previous network, that is, the early modules will affect the performance of the subsequent modules. Therefore, when selecting $K_i$, the following selection principle is followed: when $i < j$, $K_i > K_j$. In the present invention, the values of $K$ for the four stages are 256, 128, 64, and 32 respectively. That is, for deeper layers, the number of selected foreground regions is smaller to enhance the distinctiveness.

After obtaining the foreground and background features, we use an activation function and mean squared error loss to ensure the effective separation of the foreground and background. We select Hardtanh as the activation function. It does not strictly bind the output within the interval of 0 to 1, providing a certain degree of flexibility that facilitates optimization.
The Hardtanh function is an improved version of the tanh function, which is defined as shown in Eq. (\ref{hardtanh}). We use Eq. (\ref{pk}) to map the background feature $Y_{\text{max},i}^{\overline{{K_i}}}$ to a range that is not restricted by probability, so as to better remove redundant information.
\begin{equation}
  \text{Hardtanh}(x) = \frac{2 \tanh(x) - 1}{1 + \tanh^2(x)} \label{hardtanh}
\end{equation}
\begin{equation}
  P_i^{\overline{K_i}} = \text{Hardtanh}\left(Y_{\text{max},i}^{\overline{K_i}}\right) \label{pk}
\end{equation}

The final training objective is to make the predicted value of the background $P_i^{\overline{K_i}}$ tend toward -1, so as to expand the difference between the foreground and the background. In this way, the foreground and the background can be effectively segmented, providing assistance for fine-grained classification. In this process, we use the mean squared error $ \mathcal{L}_{FB} $ between the predicted value $P_i^{\overline{K_i}}$ and the target value \(- 1\) as the loss function for the \(i\)-th stage, as shown in Eq. (\ref{loss}).
\begin{equation}
  \mathcal{L}_{FB}^i = \left( P_i^{\overline{K_i}} + 1 \right)^2 \label{loss}
\end{equation}

\textbf{Architecture Variants.} We have constructed different variants of RoadFormer, including RoadFormer-T/S/B/L. The number of their parameters is similar to that of other state-of-the-art methods based on Transformer, such as Swin-T/S/B/L. The architecture specifications are listed in Table \ref{variants}. The variants differ in the number of channels $C$ and the number of blocks in each stage.

\section{Experiments}
\textbf{Dataset.} Existing autonomous driving public datasets, such as KITTI \cite{62} and Cityscape \cite{63}, focus more on the perception of traffic environments and lack annotations for road surface conditions. However, road surface information is also crucial for intelligent vehicle perception systems. The large-scale road image classification dataset RSCD \cite{64} fills this gap. It takes into account road friction levels, roughness, and materials for intelligent assisted driving. 

This dataset was originally published in \cite{65}, containing 370,000 images covering approximately 240 kilometers of road. Later, the scale of the dataset was expanded to one million. The actual driving environment of vehicles is quite complex and variable, leading to a wide variety of captured road image patterns. To address this situation, the dataset covers road images with different materials, varying years of use, and different traffic volumes under different seasons, weather, and lighting conditions.

\begin{figure}[htbp]
    \centering 
    \includegraphics[width=0.5\textwidth]{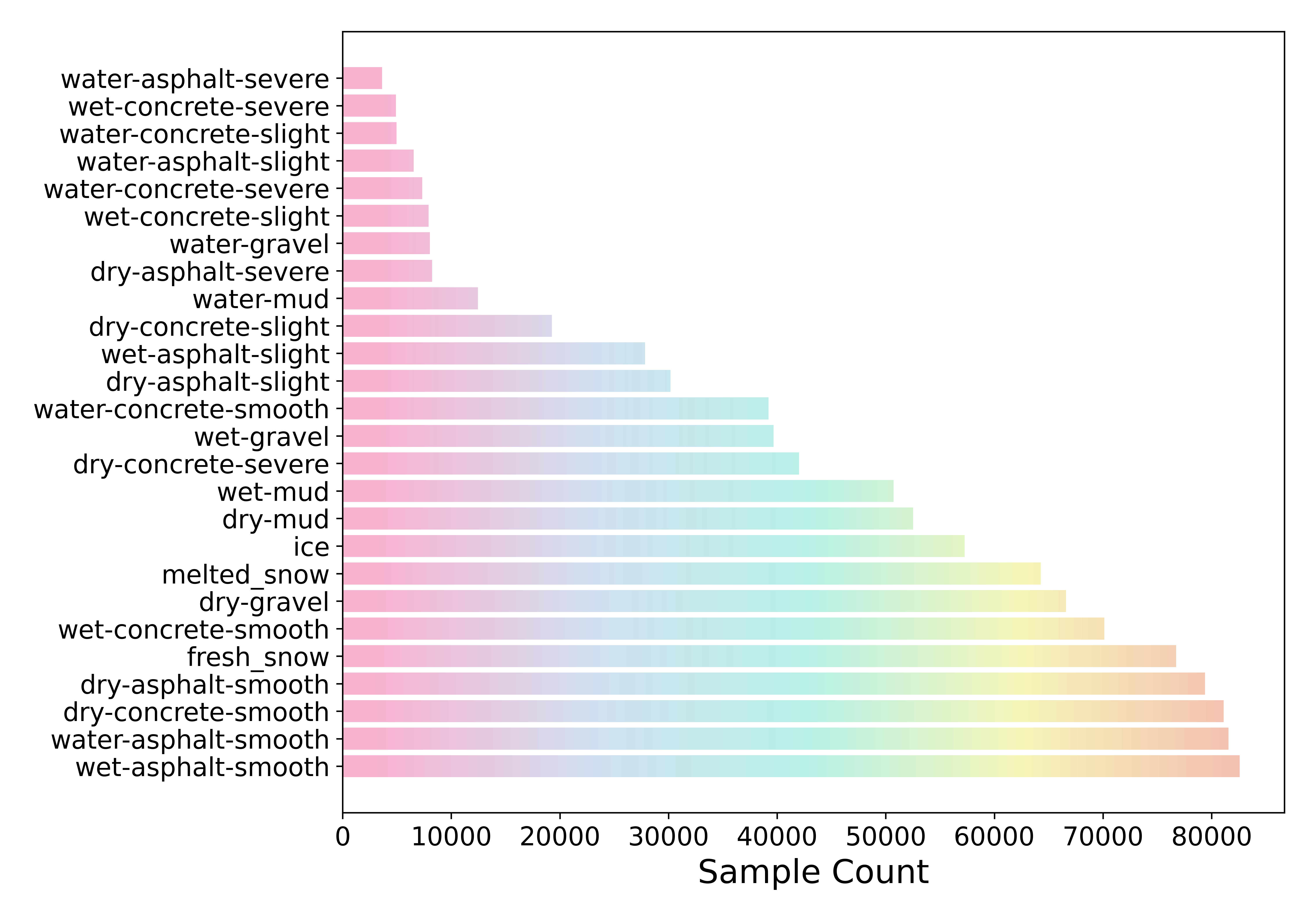}
    \caption{\textbf{RSCD:} Count of images for the 27 classes. Including but not limited to water-asphalt-severe, wet-concrete-severe, water-concrete-slight, water-asphalt-slight, etc. Each category is formed by the combination of three classification criteria: friction levels, materials, and unevenness of the road surface.}
    \label{RSCD}
\end{figure}

\begin{figure}[htbp]
    \centering 
    \includegraphics[width=0.5\textwidth]{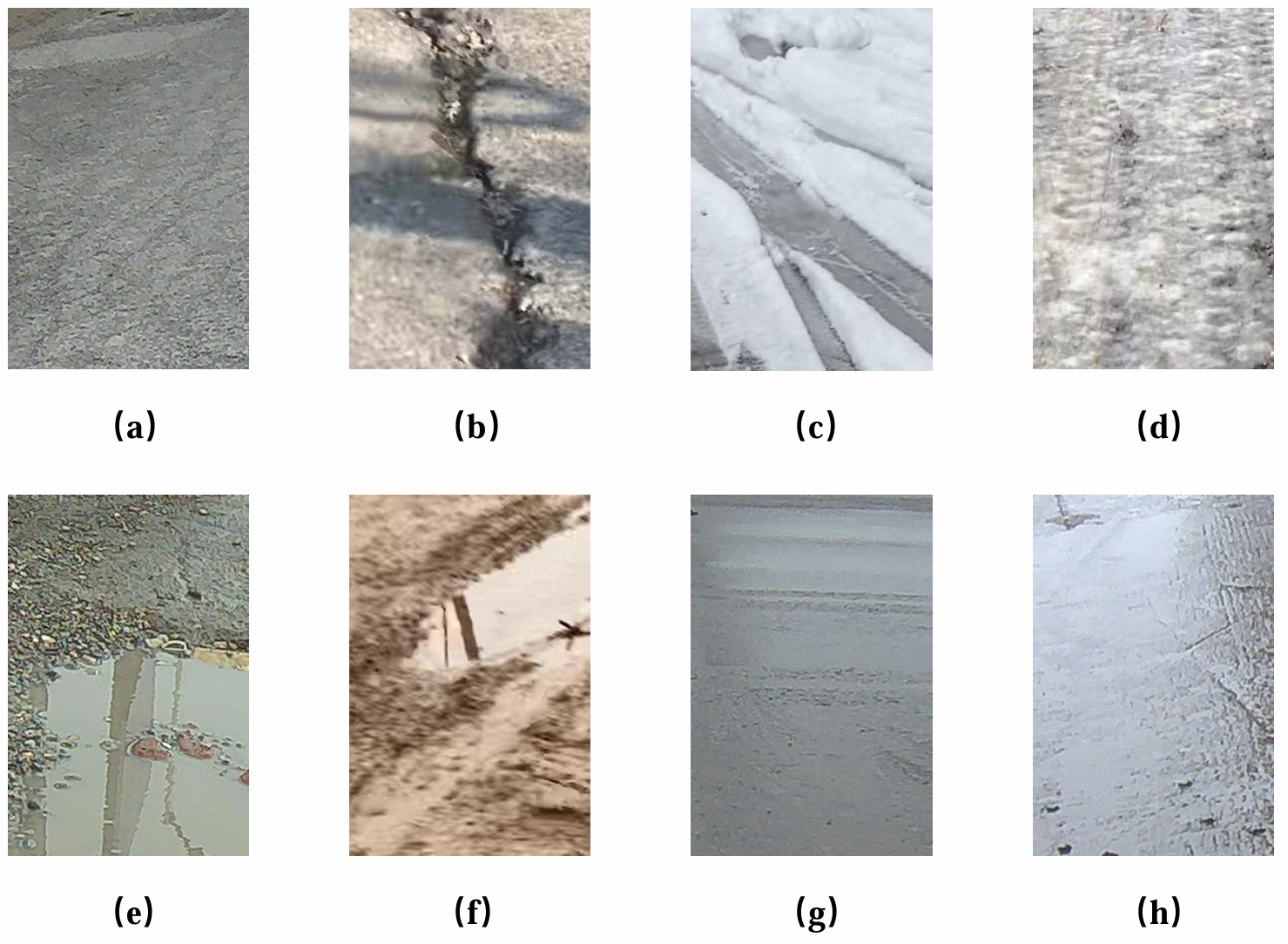}
    \caption{\textbf{Image samples of part of the classes.} (a)dry-asphalt-severe (b)dry-concrete-severe (c)fresh-snow (d)ice (e)water-gravel (f)water-mud (g)wet-asphalt-smooth (h)wet-concrete-smooth.}
    \label{sample}
\end{figure}

RSCD annotates the friction levels, materials, and unevenness of the road surface. The friction level attributes include six subclasses corresponding to different weather conditions, namely dry, wet, water, fresh snow, melted snow, and ice. Road material attributes include asphalt, concrete, mud, and gravel. Road surface unevenness is divided into smooth, slight unevenness, and severe unevenness based on the amplitude of the surface undulations. The three attributes are combined with each other to form a total of 27 combined classes, and the specific class information and some sample class diagrams are shown in Figure~\ref{RSCD} and Figure~\ref{sample}. 

\begin{figure}[htbp]
    \centering 
    \includegraphics[width=0.5\textwidth]{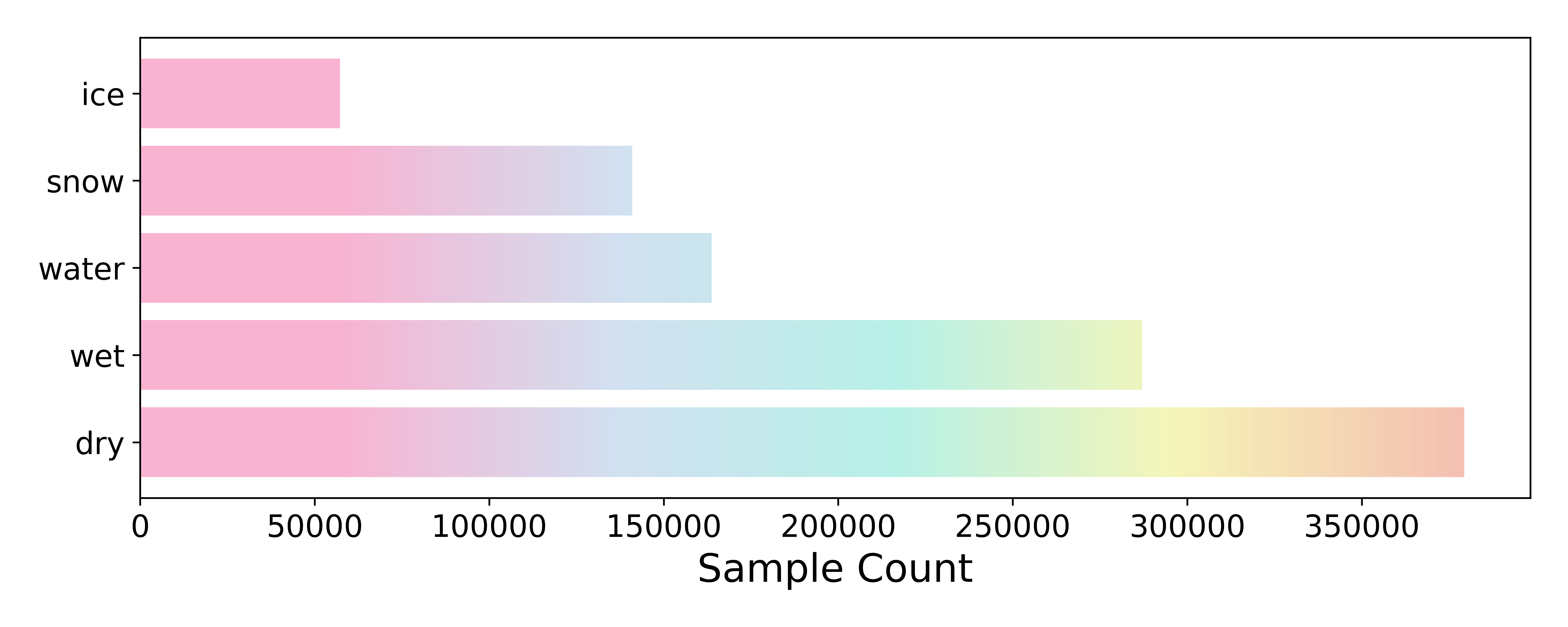}
    \caption{\textbf{simple-RSCD:} Count of images for the 5 classes. Including five categories: ice, snow, water, wet, and dry.}
    \label{simple-RSCD}
\end{figure}

Overly fine classification can indeed have better adaptability and robustness when facing diverse and complex road environments, and can deal with more actual scenario variations. However, fine-grained classification requires a large amount of detailed labeled data, which not only increases the time and cost of data preparation but may also face challenges in labeling consistency and accuracy. To achieve the same classification accuracy, fine-grained classification models are usually more complex, have more parameters, and require higher computational resources for training and inference, making them unsuitable for resource-limited autonomous driving scenarios. Therefore, we simplified the classification of RSCD. Labels such as friction conditions have a significant impact on aspects like the acceleration, braking, and steering of a vehicle, and they are key factors influencing vehicle safety. Therefore, this type of label was selected for the recombination of the dataset. We merged the dataset into five categories: dry, wet, water, snow, and ice. The sample information of each category after the dataset was reorganized is shown in Figure~\ref{simple-RSCD}. Here, the reorganized dataset is named simple-RSCD.

\textbf{Experimental Setup.} All training experiments and inference experiments are conducted on 1 RTX-4090 GPU with batch size 32. we use the AdamW optimizer with a learning rate of $5\times10^{-4}\times\frac{32}{512}$. During training, we can adjust the initial learning rate according to the batch size. This linear scaling strategy is helpful for maintaining the stability of training under different batch sizes. Other parameter settings are as follows: The weight decay is set to $0.05$, numerical stability constant $\varepsilon$ is set to $1\times10^{-8}$, momentum parameters $\beta_1$ and $\beta_2$ are set to $0.9$ and $0.999$. 
The learning rate scheduling adopts the Cosine Annealing strategy to iteratively adjust the learning rate, and the linear warmup strategy is introduced.

We use multiple image classification evaluation metrics to assess the classification performance, including Top-1 Acc, Mean Precision, Mean Recall, and Mean F1 Score. Top-1 Accuracy is one of the most commonly used evaluation metrics. It represents the proportion of samples for which the class with the highest predicted probability by the model is consistent with the true label, directly reflecting the performance of the model in single-label classification tasks. Precision measures the proportion of samples that the model predicts as positive classes and are actually positive classes. Recall, on the other hand, reflects the proportion of all samples that are actually positive classes and are correctly predicted by the model. These two metrics evaluate the classification ability of the model from different perspectives. Due to the class imbalance in the dataset, the F1 Score is introduced to balance the Precision and Recall of the model.

\definecolor{lightpink}{rgb}{0.95, 0.85, 0.85} 
\begin{table}[b]
    \centering
    \caption{Comparison Experiments between RoadFormer and other SOTA models.} 
        \renewcommand{\arraystretch}{1.2}
        \setlength{\tabcolsep}{4pt}
        \begin{tabular}{cccccc}
           \toprule
           Model & Top-1 Acc & Mean-P & Mean-R & Mean-F1 & params \\
           \midrule
           \rowcolor{lightpink}
           RoadFormer-T & 90.53 & 82.80  & 79.24 & 80.81 & 27M \\
           ConvNeXt-T\cite{61}   & 82.27 & 73.35 & 68.35 & 70.27 & 29M \\
           Swin-T\cite{34}       & 85.39 & 77.52 & 72.52 & 74.55 & 28M \\
           \midrule
           \rowcolor{lightpink}
           RoadFormer-S & 91.99 & 85.25 & 82.36 & 83.68 & 44M \\
           ConvNeXt-S\cite{61}   & 84.08 & 76.35 & 70.24 & 72.59 & 50M \\
           Swin-S\cite{34}       & 85.16 & 77.21 & 72.44 & 74.44 & 50M \\
            \midrule
           \rowcolor{lightpink}
           RoadFormer-B & 92.52 & 85.68 & 83.34 & 84.42 & 80M \\
           ConvNeXt-B\cite{61}   & 84.08 & 75.55 & 70.50 & 72.48 & 89M \\
           Swin-B\cite{34}       & 85.68 & 77.69 & 73.24 & 75.08 & 88M \\
           ViT-B\cite{33}        & 86.83 & 78.31 & 74.99 & 76.38 & 86M \\
           \midrule
           \rowcolor{lightpink}
           RoadFormer-L & 92.86 & 86.17 & 83.95 & 84.99 & 206M \\
           ConvNeXt-L\cite{61}   & 84.88 & 76.65 & 72.36 & 74.20 & 198M \\
           Swin-L\cite{34}       & 87.38 & 80.47 & 76.03 & 77.91 & 197M \\
           ViT-L\cite{33}        & 88.47 & 79.99 & 77.51 & 78.58 & 307M \\
           \bottomrule
        \end{tabular} \label{compare}
\end{table}

\textbf{Comparative experiment.}

We conducted image classification experiments on RSCD. For a fair comparison, all models were trained for 40 epochs on the same device, and the resolution of all input images was adjusted to $224\times 224$. As shown in Table \ref{compare}, compared to the SOTA method, our method is ahead of the curve on all metrics.     

Taking the model with the base size as an example, in terms of four evaluation metrics, namely Top-1 Accuracy, Mean Precision, Mean Recall, and Mean F1 Score, compared with the excellent traditional convolutional method ConvNeXt, our method has improved by 8.44\%, 10.13\%, 12.84\%, and 11.94\% respectively. Compared with the efficient hierarchical attention mechanism Swin-Transformer, our method has improved by 6.84\%, 7.99\%, 10.01\%, and 9.34\% respectively. Compared with the pioneering global attention model ViT, our method has improved by 5.69\%, 7.37\%, 8.35\%, and 8.04\% respectively. Moreover, compared with the above three models, the number of parameters of our model has decreased by 11.25\%, 9.1\%, and 6.98\% respectively. This indicates that the RoadFormer proposed by us is an effective and promising paradigm.

\textbf{Stacking Structure.}
To address the unique feature extraction problem in fine-grained pavement classification, in this section, we explore different ways of stacking local modules and global modules. As shown in Table \ref{stack}, the number of satges is kept constant at 4 for each stacking method and the number of parameters is kept in the same order of magnitude.  In the table, $L$ represents the local stage, which only includes the Conv Block responsible for local feature extraction, $G$ represents the global stage, which only includes the Trans Block responsible for global feature extraction, and $M$ represents the local-global mixed stage.

\begin{table}[htbp]
    \centering
    \caption{Comparison Experiments between different Stacking Structures.} 
        \renewcommand{\arraystretch}{1.2}
        \begin{tabular}{ccccc}
           \toprule
           Model & Top-1 Acc & Mean-P & Mean-R & Mean-F1 \\
           \midrule
           \rowcolor{lightpink}
           LMMG & 92.52 & 85.68 & 83.34 & 84.42 \\
           LMMM & 92.33 & 85.41 & 82.70 & 83.96 \\
           LMGG & 92.31 & 85.44 & 82.95 & 84.10 \\
           \bottomrule
        \end{tabular} \label{stack}
\end{table}

\begin{figure*}[t]
    \centering
    \includegraphics[width=\textwidth]{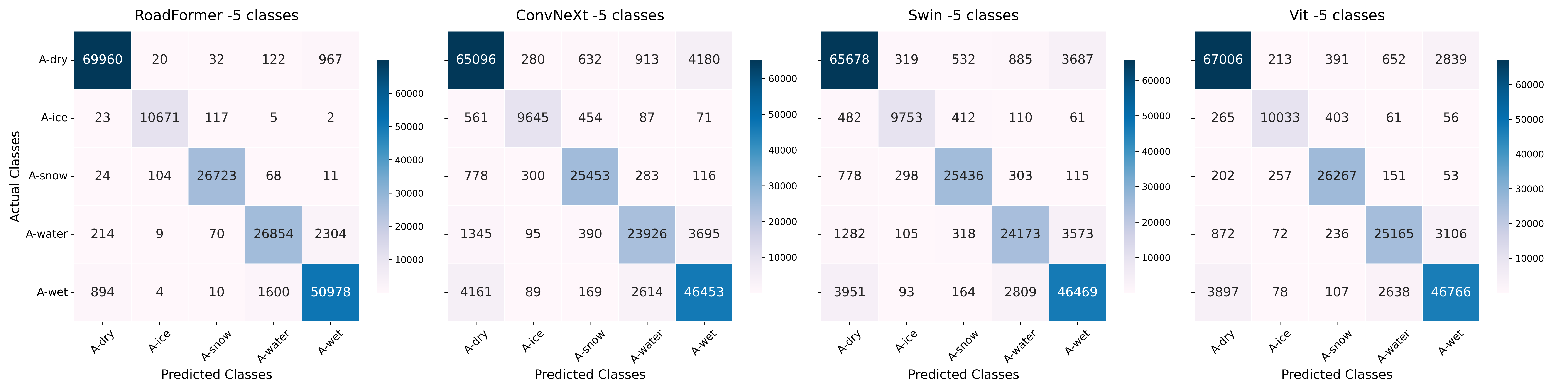}
    \caption{\textbf{The confusion matrix on Simple RSCD test set.} Shows the classification performance of four models : RoadFormer, ConvNeXt, Swin, and Vit on Simple RSCD.}
    \label{5class}
\end{figure*}

\textbf{Is the FBM effective?}

To verify the universality of the introduced FBM in the fine-grained pavement classification task, we added this module to ConvNeXt, Swin-Transformer, and ViT respectively. The results are shown in Table \ref{fbm}. As we can observe that the FBM also demonstrates certain effectiveness on other models. There is an improvement of approximately 0.1\% in the Top-1 Acc metric for all three models, and other metrics also show some increase to a certain extent.

\begin{table}[htbp]
    \centering
    \caption{The performance of FBM on different models.} 
        \renewcommand{\arraystretch}{1.2}
        \begin{tabular}{ccccc}
           \toprule
           Model & Top-1 Acc & Mean-P & Mean-R & Mean-F1 \\
           \midrule
           ConvNeXt\cite{61}       & 83.69 & 75.66 & 69.37 & 71.75 \\
           \rowcolor{lightpink}
           ConvNeXt+FBM & 83.79 & 76.46 & 69.95 & 72.39 \\
           \midrule
           Swin\cite{34}           & 82.91 & 73.60 & 68.49 & 70.52 \\
           \rowcolor{lightpink}
           Swin+FBM     & 83.01 & 74.08 & 68.38 & 70.53 \\
           \midrule
           ViT\cite{33}            & 84.19 & 74.44 & 70.63 & 72.12 \\
           \rowcolor{lightpink}
           ViT+FBM      & 84.28 & 74.57 & 70.67 & 72.23 \\
           \bottomrule
        \end{tabular} \label{fbm}
\end{table}

\textbf{Ablation Experiment.}

In this section, we present a set of ablation experiments to verify the effectiveness of the FBM and the proposed novel stacking method. As shown in Table \ref{ablation}, when neither the FBM nor the hybrid stacking structure is adopted, the Top-1 Acc of the model is 91.9\%, which is already a good benchmark performance. When the FBM is adopted alone, the performance is improved to 92.23\%, indicating that the FBM has a certain positive impact on the model performance. When the hybrid stacking structure is adopted alone, the performance is improved to 92.34\%, which means that the novel stacking method we proposed is indeed effective. When both the FBM and the hybrid stacking structure are adopted simultaneously, the performance reaches 92.52\%, which is higher than the benchmark performance and the performance when they are used separately, demonstrating that the two work well in tandem.

\begin{table}[htbp]
    \centering
    \caption{Ablation Experiment,  F represents Front Background Module, S represents Stacking Structure.} 
        \renewcommand{\arraystretch}{1.2}
        \begin{tabular}{ccccc}
           \toprule
           Model & Top-1 Acc & Mean-P & Mean-R & Mean-F1 \\
           \midrule
           Without F\&S & 91.90 & 84.86 & 81.53 & 83.04 \\
           Without F    & 92.34 & 85.67 & 83.14 & 84.31 \\
           Without S    & 92.23 & 85.21 & 82.61 & 83.80 \\
           \rowcolor{lightpink}
           RoadFormer-B & 92.52 & 85.68 & 83.34 & 84.42 \\
           \bottomrule
        \end{tabular} \label{ablation}
\end{table}

\textbf{Experiments on a simple dataset.}

\begin{figure}[t]
    \centering
    \includegraphics[width=0.5\textwidth]{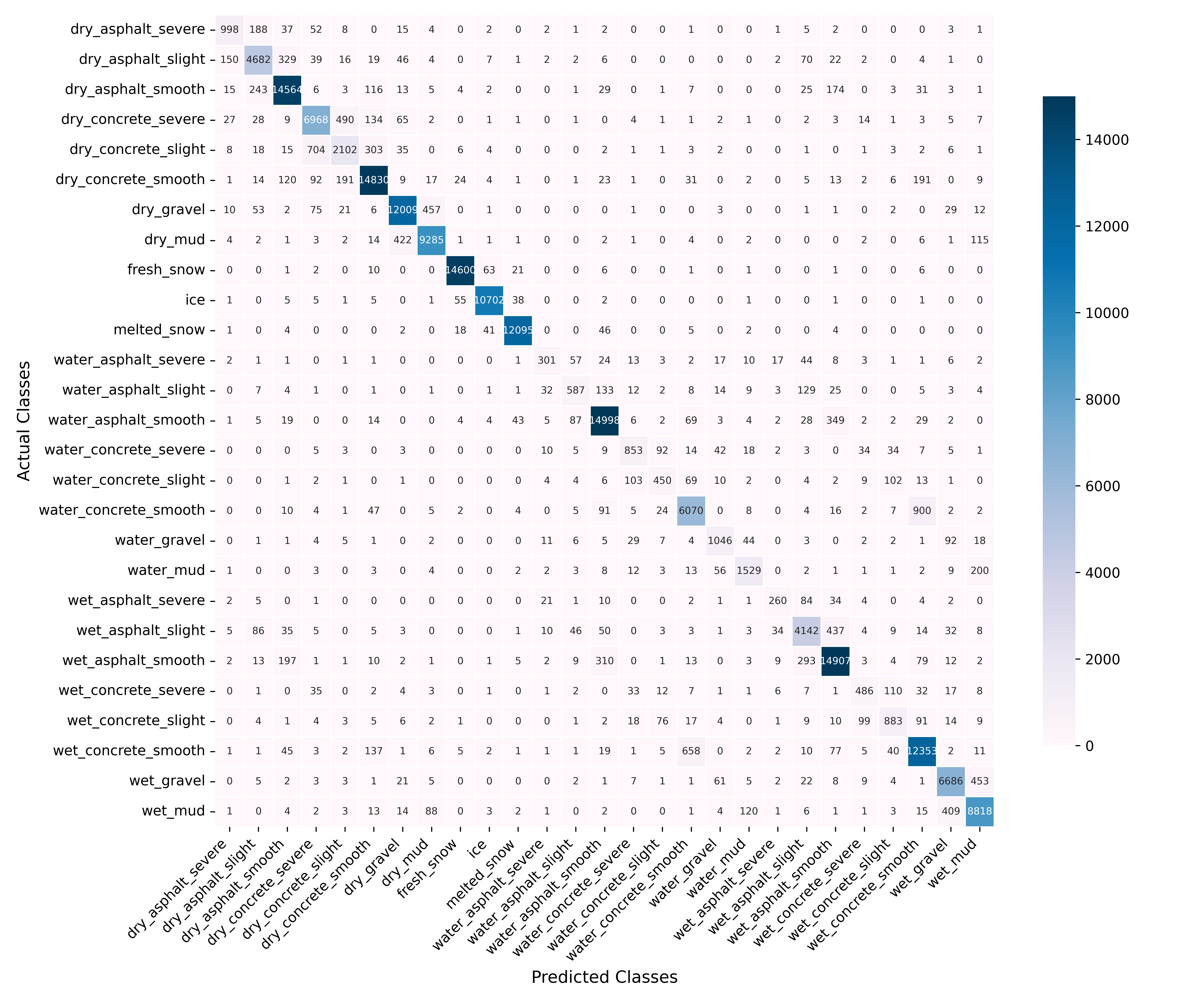}
    \caption{\textbf{The confusion matrix on RSCD test set.} Shows the classification performance of RoadFormer on RSCD.}
    \label{27class}
\end{figure}

We conducted experiments on a 5-class dataset simplified from the RSCD dataset, aiming to explore the model's adaptability to different classification granularities. In actual autonomous driving scenarios, the choice of the classification granularity of the road surface often depends on specific requirements. Performing experiments on two datasets provides a basis for model selection and optimization for different application scenarios. As shown in Table \ref{simple}, it can be found that the performance of RoadFormer on simple-RSCD is significantly improved, showing its powerful ability.

\begin{table}[htbp]
    \centering
    \caption{The performance of RoadFormer on Simple-RSCD.}  
        \renewcommand{\arraystretch}{1.2}
        \begin{tabular}{ccccc}
           \toprule
           Model & Top-1 Acc & Mean-P & Mean-R & Mean-F1 \\
           \midrule
           RoadFormer-T & 96.09 & 96.36 & 96.10 & 96.23 \\
           RoadFormer-S & 96.15 & 96.44 & 96.16 & 96.29 \\
           RoadFormer-B & 96.50 & 96.72 & 96.50 & 96.61 \\
           RoadFormer-L & 96.63 & 96.86 & 96.62 & 96.73 \\
           \bottomrule
        \end{tabular} \label{simple}
\end{table}

\textbf{Confusion Matrix Analysis.}

The confusion matrix presents the corresponding relationship between the model's prediction results and the actual labels in the form of a matrix, which can intuitively reflect the classification performance of the model for different categories. Specifically, the confusion matrix intuitively reveals the model's classification accuracy and error distribution by counting the number of true positives (TP), false negatives (FN), false positives (FP), and true negatives (TN) for each category. Based on the confusion matrix, key metrics such as precision (Precision = TP / (TP + FP)), recall (Recall = TP / (TP + FN)), and specificity (Specificity = TN / (TN + FP)) can be further calculated to comprehensively evaluate the model's performance across different types of errors. Figure~\ref{5class} and Figure~\ref{27class} show the confusion matrices obtained from experiments conducted on simple-RSCD and RSCD using the four models: RoadFormer-B, ConvNeXt-B, Swin-B, and ViT-B. Analysis reveals that, whether on simple-RSCD or RSCD, our model outperforms the other three models, with the number of correctly classified samples in each category exceeding that of the other three.

Through an in-depth analysis of the confusion matrix, we found that the misclassification of the model between the "dry-wet" label pair and the "wet-water" label pair is relatively prominent. This phenomenon may stem from the fact that there is still room for improvement in model performance, as well as subjective interference during the labeling process of the dataset. Unlike other fine-grained classification tasks, the definition of pavement categories itself has subjective judgment differences, which further exacerbates the model's confusion.

\section{Conclusion}

In this paper, we propose a network architecture for fine-grained road surface classification in autonomous driving scenarios. In view of the special nature of the fine-grained classification task, a foreground background module and a novel stacking structure of local-global feature extraction modules are introduced. The Top-1 accuracy of our method on the RSCD test set reaches 92.52\%, and on the simple-RSCD test set, it even reaches 96.50\%. Compared to existing SOTA methods across all model sizes, classification accuracy improved by 5.69\% to 12.84\%. Finally, we identify the defects and causes of the current method. In future research, attention should be focused on the optimization of the model architecture and the consistency of data annotation. It is necessary to explore more advanced model architectures to improve the feature extraction ability. At the same time, clear annotation specifications should be formulated, and annotators should be trained to ensure the consistency and accuracy of annotation. A multiple annotation mechanism should be introduced to review controversial samples, reduce subjective interference, and improve data quality. Through above measures, we expect to effectively alleviate the model confusion problem in future research and enhance the accuracy and reliability of road surface category recognition. In conclusion, our work provides valuable references for further research related to road perception.

\bibliographystyle{ieeetr} 
\bibliography{references} 

\end{document}